\documentclass[lettersize,journal]{IEEEtran}
\usepackage{amsmath,amsfonts}
\usepackage[noend]{algpseudocode}
\usepackage{booktabs}
\usepackage{indentfirst}
\usepackage{threeparttable}
\usepackage{algorithmicx}
\usepackage{algorithm}
\usepackage{amsmath}
\usepackage{array}
\usepackage{multirow}
\usepackage{textcomp}
\usepackage{stfloats}
\usepackage{url}
\usepackage{tabularx}
\usepackage{verbatim}
\usepackage{graphicx}
{}
\usepackage{float}
\usepackage{booktabs}
\usepackage{subfigure}
\usepackage{cite}
\usepackage{colortbl}
\usepackage{CJKutf8}
\usepackage{hyperref}
\usepackage{cleveref}
\definecolor{mygreen}{rgb}{0.62,0.92,0.72}

\usepackage{booktabs} 
\usepackage{multirow} 
\usepackage{array}    

\usepackage{amssymb}  
\usepackage{booktabs}  

\usepackage{pifont} 
\usepackage{xcolor} 
\usepackage{colortbl} 
\usepackage{orcidlink}
\newcommand{\funcapply}{\operatorname{apply}}
\definecolor{textgreen}{rgb}{0.0, 0.5, 0.0}
\begin{document}
\begin{CJK}{UTF8}{gbsn}

\title{Towards Efficient 3D Object Detection for Vehicle-Infrastructure Collaboration via Risk-Intent Selection}

\author {
Li Wang\href{https://orcid.org/0000-0002-9325-2391}{\orcidlink{0000-0002-9325-2391}},
Boqi Li,
Hang Chen,
Xingjian Wu, 
Yichen Wang,
Jiewen Tan,
Xinyu Zhang\href{https://orcid.org/0000-0003-0034-9037}{\orcidlink{0000-0003-0034-9037}},
and Huaping Liu\href{https://orcid.org/0000-0002-4042-6044}{\orcidlink{0000-0002-4042-6044}}, \textit{Fellow, IEEE}

\thanks{This work was supported by the National Natural Science Foundation of China under Grant No. 52502496, U22B2052 and the Natural Science Foundation of Chongqing, China under Grant No. CSTB2025NSCQ-GPX0413, and the National High Technology Research and Development Program of China under Grant No. 2020YFC1512501. \textit{(Corresponding author: Boqi Li.)}}
\thanks{Li Wang is with School of Mechanical Engineering, Beijing Institute of Technology, Beijing 100081, China, Chongqing Innovation Center, Beijing Institute of Technology, Chongqing 401120, China, and Hubei Key Laboratory of Intelligent Technologies for Transportation Internet of Things, Wuhan University of Technology, Wuhan 430070, China (e-mail: wangli\_bit@bit.edu.cn).}
\thanks{Boqi Li is with Civil and Environmental Engineering Department, University of Michigan, Ann Arbor 48109, USA (e-mail: boqili@umich.edu).}
\thanks{Hang Chen is with the School of Software Engineering, Sun Yat-Sen University, Guangzhou 510000, China (e-mail:  chenh797@mail2.sysu.edu.cn).}
\thanks{Xingjian Wu is with the State Key Laboratory of Robotics and Systems, Harbin Institute of Technology, Harbin 150001, China (e-mail: 24S008084@stu.hit.edu.cn).}
\thanks{Yichen Wang is with School of Astronautics, Harbin Institute of Technology, Harbin, 150006, China (e-mail: 22S136042@stu.hit.edu.cn).}
\thanks{Jiewen Tan is with the Henry Samueli School of Engineering, University of California, Irvine, CA 92697, USA (e-mail: jiewent1@uci.edu).}
\thanks{Xinyu Zhang is with the School of Vehicle and Mobility, Tsinghua University, Beijing 100084, China (e-mail: xyzhang@ts-inghua.edu.cn).}
\thanks{Huaping Liu is with the State Key Laboratory of Intelligent Technology and Systems, and the Department of Computer Science and Technology, Tsinghua University, Beijing 100084, China (e-mail: hpliu@tsinghua.edu.cn).}
}

\maketitle

\begin{abstract}
Vehicle-Infrastructure Collaborative Perception (VICP) is pivotal for resolving occlusion in autonomous driving, yet the trade-off between communication bandwidth and feature redundancy remains a critical bottleneck. While intermediate fusion mitigates data volume compared to raw sharing, existing frameworks typically rely on spatial compression or static confidence maps, which inefficiently transmit spatially redundant features from non-critical background regions. To address this, we propose Risk-intent Selective detection (RiSe), an interaction-aware framework that shifts the paradigm from identifying visible regions to prioritizing risk-critical ones. Specifically, we introduce a Potential Field-Trajectory Correlation Model (PTCM) grounded in potential field theory to quantitatively assess kinematic risks. Complementing this, an Intention-Driven Area Prediction Module (IDAPM) leverages ego-motion priors to proactively predict and filter key Bird's-Eye-View (BEV) areas essential for decision-making. By integrating these components, RiSe implements a semantic-selective fusion scheme that transmits high-fidelity features only from high-interaction regions, effectively acting as a feature denoiser. Extensive experiments on the DeepAccident dataset demonstrate that our method reduces communication volume to 0.71\% of full feature sharing while maintaining state-of-the-art detection accuracy, establishing a competitive Pareto frontier between bandwidth efficiency and perception performance.
\end{abstract}

\section{Introduction}

\label{sec:Introduction}

\IEEEPARstart
{I}{n} recent years, autonomous driving perception technology\cite{zhu2024accurate,zhang2020detection,cheng2024ef,le2024multilevel,li2023mvmm} has experienced rapid development. The recent emergence of Bird's-Eye-View (BEV) representation learning\cite{li2022bevformer,liu2022bevfusion,philion2020lift} has become a prevailing paradigm, providing a unified and spatial-aware feature space for multi-sensor fusion and 3D perception tasks\cite{liu2022petr,wang2021detr3d}. However, limited by sensor field of view and occlusions, independent perception struggles to acquire global scene information, posing safety risks, especially in complex intersections and oncoming traffic scenarios. Collaborative perception offers a new solution by extending the perception range and improving robustness through vehicle-infrastructure cooperation, which is crucial for achieving safer and more reliable autonomous driving\cite{huang2025vehicle}. 

\begin{figure*}
    \centering
    \includegraphics[width=0.87\linewidth]{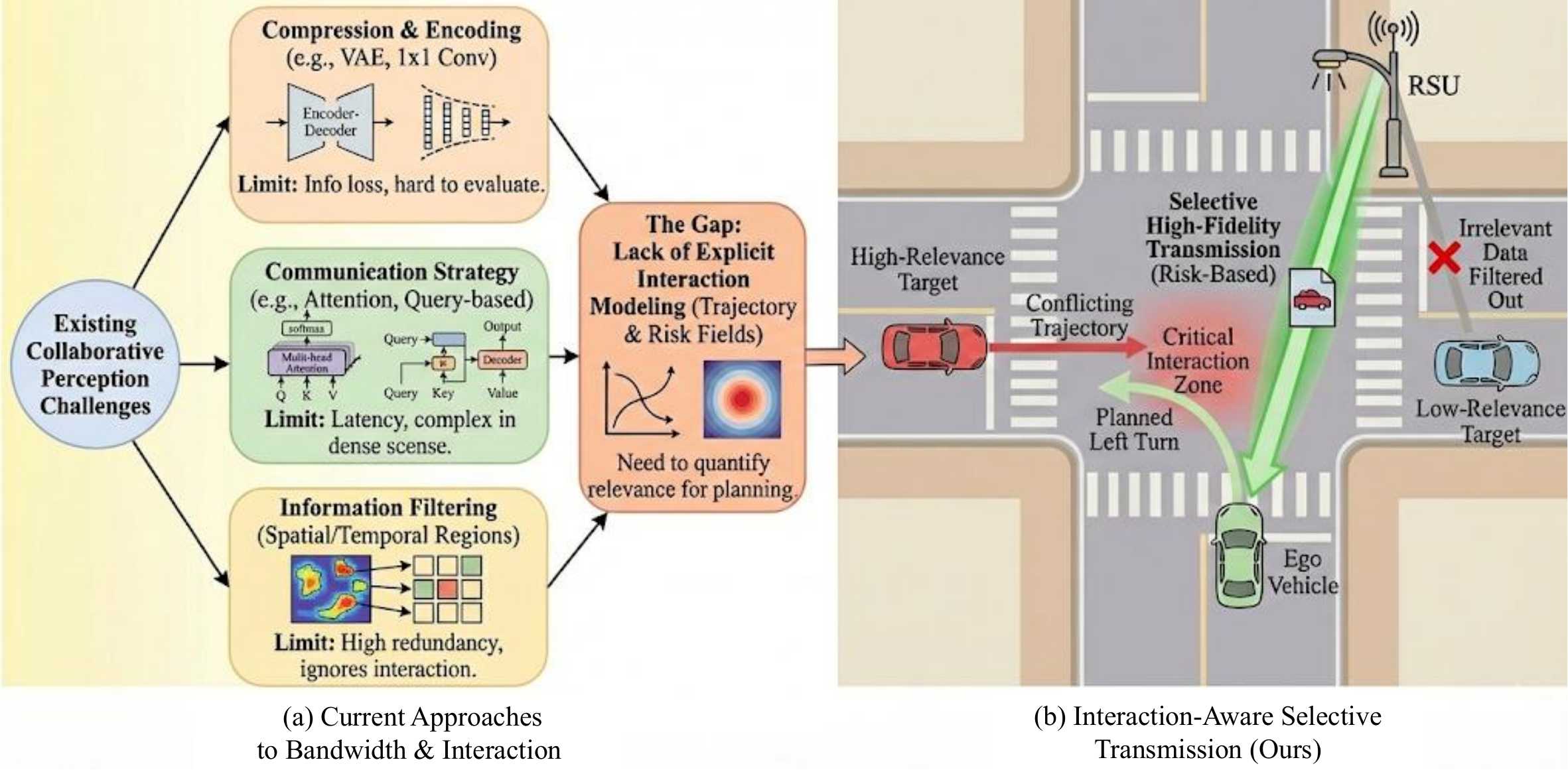}
    \caption{The core motivation and evolution of the proposed RiSe framework. (a) Evolution of collaborative perception: moving from bandwidth-limited compression and static filtering to our proposed interaction-driven paradigm that utilizes risk fields. (b) Motivation scenario: leveraging the ego-vehicle's driving intent to identify the Critical Interaction Zone, allowing the infrastructure to prioritize high-risk targets (red) over irrelevant ones (blue) to optimize bandwidth efficiency.}
    \label{fig:placeholder}
\end{figure*}

As illustrated in Fig.~\ref{fig:placeholder}(b), the utility of perceptual information is inherently non-uniform. While the oncoming vehicle (red) poses a high interaction risk to the ego vehicle's left-turn plan, the departing vehicle (blue) is spatially irrelevant. Although Vehicle-Infrastructure Collaborative (VIC) perception extends the sensing range, conventional methods often incur high communication overhead by broadcasting these redundant features. To address this, we propose RiSe, which leverages driving intent to identify the Critical Interaction Zone, allowing the infrastructure to selectively transmit high-fidelity features only where safety is at stake. This aligns with the broader trend of developing planning-oriented \cite{hu2022uniad} and end-to-end systems that tightly couple perception, prediction, and decision-making. Based on information type and collaboration stage, collaborative perception solutions can be categorized into early, mid-stage, and late fusion \cite{liu2023vehicletoeverythingautonomousdrivingsurvey}. Among these, mid-stage fusion has become the mainstream solution due to its lower bandwidth requirements by transmitting feature-level information.

However, existing collaborative perception solutions, especially mid-stage fusion, still face several challenges. First, despite the reduced bandwidth requirements of feature-level information, information redundancy remains a problem. To further reduce bandwidth, some studies employ lossy compression methods, such as the encoder-decoder information sharing paradigm using 1x1 convolution for feature compression and reconstruction \cite{li2022learningdistilledcollaborationgraph,xu2022opv2vopenbenchmarkdataset,wang2023corecooperativereconstructionmultiagent,xu2022v2xvitvehicletoeverythingcooperativeperception}. Many of these contemporary methods operate on BEV features\cite{li2022bevformer,liu2022bevfusion}, leveraging its advantages for spatial reasoning and multi-agent fusion. This approach is susceptible to perception feature noise, packet loss, and numerical truncation in practical applications, leading to system instability and even reduced perception accuracy.\cite{xu2025v2x} Other methods attempt to transmit local information of raw features \cite{hu2022where2commcommunicationefficientcollaborativeperception,10265751,10.1145/3581783.3611699,wang2023umcunifiedbandwidthefficientmultiresolution,yang2023spatiotemporaldomainawarenessmultiagent,hu2023collaborationhelpscameraovertake,wei2023asynchronyrobustcollaborativeperceptionbirds,10.1145/3503161.3548197}, utilizing an effective information filtering mechanism to transmit valid perception information with minimal loss. However, in complex traffic scenarios, especially with an increasing number of cooperating agents, effectively selecting and transmitting information from all potential perceived target regions while preventing exponential growth of redundant information remains a significant challenge. Second, collaborative perception scenarios contain rich vehicle interaction information that is unavailable to independent perception. Advances in perception prediction enable us to predict the future driving states of target vehicles, which can better serve ego-vehicle path planning and decision-making \cite{gao2020vectornetencodinghdmaps,gilles2021homeheatmapoutputfuture,zhao2020tnttargetdriventrajectoryprediction,9157523,Fang_2020_CVPR,hu2021FIERYfutureinstanceprediction}. Existing efforts have explored incorporating planning states or other interaction cues to guide the exchange of collaborative perception information. However, these methods largely overlook the critical role of inter-vehicle interactions—such as relative spatial proximity and kinematic dependencies—in determining perceptual relevance. Effectively modeling such interactions allows the system to identify safety-critical targets that may influence the ego vehicle’s planning and decision-making processes, thereby enabling adaptive allocation of communication resources and enhancing situational awareness in dynamic environments\cite{dong2025task}.

To this end, we propose RiSe (Risk-intent Selective detection), a risk-aware collaborative perception framework. The main contributions are summarized as follows:
\begin{itemize}
\item We propose a vehicle interaction assessment model integrating trajectory interaction and risk fields. By quantifying kinematic risks, this model identifies targets strictly relevant to the ego-vehicle's driving task, guiding precise feature selection.
\item We develop a driving area prediction model that fuses ego-motion priors with trajectory prediction. Unlike standard heatmap approaches, this module proactively identifies high-interaction regions within the infrastructure's perception range.
\item We design an interaction-aware fusion paradigm that optimizes the bandwidth-accuracy trade-off. By transmitting BEV features solely from predicted high-risk zones, we minimize data redundancy while ensuring the detection of safety-critical targets.
\item We introduce an interaction-aware metric to assess detection on critical objects. Experiments demonstrate that RiSe achieves competitive accuracy while reducing communication volume to 0.71\% of full transmission, significantly outperforming conventional compression methods.
\end{itemize}

In summary, the core insight of this study is that collaborative perception must be selective, prioritizing information strictly based on its relevance to driving safety. We demonstrate that by quantitatively modeling vehicle interactions and integrating driving intent, we can establish a highly efficient collaboration paradigm. This work advances planning-oriented collaborative perception, ensuring that perception resources are allocated based on interaction criticality rather than mere visibility.

\section{related work}
\label{sec:RelatedWork}
\subsection{Communication mechanism among collaborative perception agents}
Collaborative perception predominantly utilizes intermediate fusion, yet bandwidth remains a critical bottleneck. 
To mitigate this, compression-based methods employ encoder-decoder networks~\cite{li2022learningdistilledcollaborationgraph,xu2022opv2vopenbenchmarkdataset,wang2023corecooperativereconstructionmultiagent,xu2022v2xvitvehicletoeverythingcooperativeperception} or Variational Autoencoders (VAE)~\cite{wang202022vnetvehicletovehiclecommunicationjoint} to reduce data volume. 
However, these approaches often incur information loss, degrading performance on small objects.
Alternatively, adaptive communication strategies focus on optimizing \textit{when} and \textit{who} to communicate~\cite{zaki2025quality}. 
For instance, attention-based mechanisms~\cite{liu2020who2comcollaborativeperceptionlearnable,liu2020when2commultiagentperceptioncommunication} and query-based frameworks like QUEST~\cite{fan2024questquerystreampractical} utilize graph structures or cross-agent queries to select informative features~\cite{zhao2025multiagent}.
While effective, these methods typically focus on reducing channel redundancy rather than assessing the semantic value of the transmitted data relative to the driving task.

Spatial-temporal filtering offers a more granular approach to bandwidth efficiency. 
Seminal works like Where2Comm~\cite{hu2022where2commcommunicationefficientcollaborativeperception} employ spatial confidence maps to selectively share features near perceived objects. 
Subsequent studies have extended this by incorporating channel-wise selection~\cite{10265751}, multi-resolution mechanisms~\cite{wang2023umcunifiedbandwidthefficientmultiresolution}, and temporal context aggregation~\cite{10.1145/3581783.3611699,yang2023spatiotemporaldomainawarenessmultiagent}. 
Recent advances further enhance feature quality via alignment~\cite{contrastalign2024} or self-distillation~\cite{jiang2024fsd_bev, zhou2025ei}.
However, a common limitation is that feature selection is often driven by "visibility" or static confidence, leading to redundancy in dense traffic.
While some recent works incorporate planning knowledge~\cite{10752404}, they often lack a quantitative evaluation of interaction levels. 
To address this, we propose an interaction assessment model based on risk fields, leveraging recent insights into heterogeneous V2X interactions~\cite{lin2025roadside,zheng2025surrogate,zhang2025near,zha2025heterogeneous} to prioritize targets that strictly influence the ego-vehicle's safety.

This model identifies critical targets that may influence the ego-vehicle's driving task and guides the precise selection of collaborative perception feature information. To visually contextualize this contribution, Fig.~\ref{fig:placeholder}(a) illustrates the paradigm shift from conventional bandwidth-reduction techniques to our proposed interaction-driven framework. Unlike static filtering, our approach leverages trajectory priors and risk assessment to dynamically allocate bandwidth to safety-critical regions.
\begin{table*}[]
\centering
\caption{\textbf{Qualitative comparison with state-of-the-art collaborative perception methods.} Unlike existing approaches that rely on compression or confidence-based filtering, RiSe is the first to explicitly integrate interaction risk modeling and driving intent guidance to minimize redundancy.}
\label{tab:related_work_comparison}
\footnotesize
\renewcommand{\arraystretch}{1.3}
\setlength{\tabcolsep}{12pt}
\begin{tabular}{@{}l|l|cc|l@{}}
\toprule
\textbf{Method} & \textbf{Core Mechanism} & \textbf{Interaction-Aware} & \textbf{Intent-Driven} & \textbf{Primary Limitation} \\
\midrule
V2VNet \cite{wang202022vnetvehicletovehiclecommunicationjoint} & VAE Compression & \textcolor{gray}{\ding{55}} & \textcolor{gray}{\ding{55}} & Lossy compression degrades small objects \\
V2X-ViT \cite{xu2022v2xvitvehicletoeverythingcooperativeperception} & Attention Masking & \ding{51} (Implicit) & \textcolor{gray}{\ding{55}} & High computational overhead \\
Where2Comm \cite{hu2022where2commcommunicationefficientcollaborativeperception} & Confidence Map & \textcolor{gray}{\ding{55}} & \textcolor{gray}{\ding{55}} & Lacks explicit risk modeling \\
QUEST \cite{fan2024questquerystreampractical} & Query Stream & \textcolor{gray}{\ding{55}} & \textcolor{gray}{\ding{55}} & Limited to query-based interactions \\
\midrule
\textbf{RiSe (Ours)} & \textbf{Risk-Aware Filtering} & \textbf{\ding{51} (Explicit)} & \textbf{\ding{51} (Active)} & \textbf{Requires ego-motion prior} \\
\bottomrule
\end{tabular}
\end{table*}
\subsection{Trajectory Prediction and Interaction Risk Assessment}
Trajectory prediction is pivotal for understanding scene dynamics. 
Vision-based collaborative perception often predicts future motion states in BEV using heatmaps~\cite{hu2023collaborationhelpscameraovertake,wei2023asynchronyrobustcollaborativeperceptionbirds}, a paradigm advanced by FIERY~\cite{gilles2021homeheatmapoutputfuture} and its successors~\cite{feng2024bevspread,li2023bevsuper}. 
While independent perception struggles with occlusion, collaborative systems can leverage infrastructure to model interactions more effectively~\cite{author2024emiff}.
Regarding interaction modeling, traditional approaches rely on data-driven methods or potential field theories to quantify mutual influence.
Notably, the concept of a "driving safety field"~\cite{6957815} utilizes field theory to model human-vehicle-road interactions and predict dynamic risks.
Inspired by this, we propose to integrate risk fields into the collaborative perception loop. 
Instead of merely predicting trajectories, we utilize kinematic risk quantification to serve as a semantic filter, identifying and transmitting only those targets that pose a potential threat to the ego-vehicle's planned path. Existing approaches primarily focus on bandwidth reduction via data compression or confidence-based filtering, but inherently lack mechanisms to assess the \textit{semantic criticality} of targets. Specifically, most methods fail to explicitly model interaction risks or leverage the ego-vehicle's driving intent to guide feature transmission. RiSe addresses this gap by introducing a risk-aware paradigm that actively prioritizes information based on its relevance to driving safety. A qualitative comparison between our proposed framework and state-of-the-art collaborative perception methods is presented in Table~\ref{tab:related_work_comparison}. As illustrated in the comparison, while existing approaches primarily focus on bandwidth reduction via data compression (e.g., V2VNet~\cite{wang202022vnetvehicletovehiclecommunicationjoint}) or confidence-based filtering (e.g., Where2Comm~\cite{hu2022where2commcommunicationefficientcollaborativeperception}), they inherently lack mechanisms to assess the \textit{semantic criticality} of targets. Specifically, most methods fail to explicitly model interaction risks or leverage the ego-vehicle's driving intent to guide feature transmission. RiSe addresses this gap by introducing a risk-aware paradigm that actively prioritizes information based on its relevance to driving safety.
\begin{figure*}[!t]
	\vspace{-5mm}
	\centering
	\includegraphics[width=0.90\linewidth]{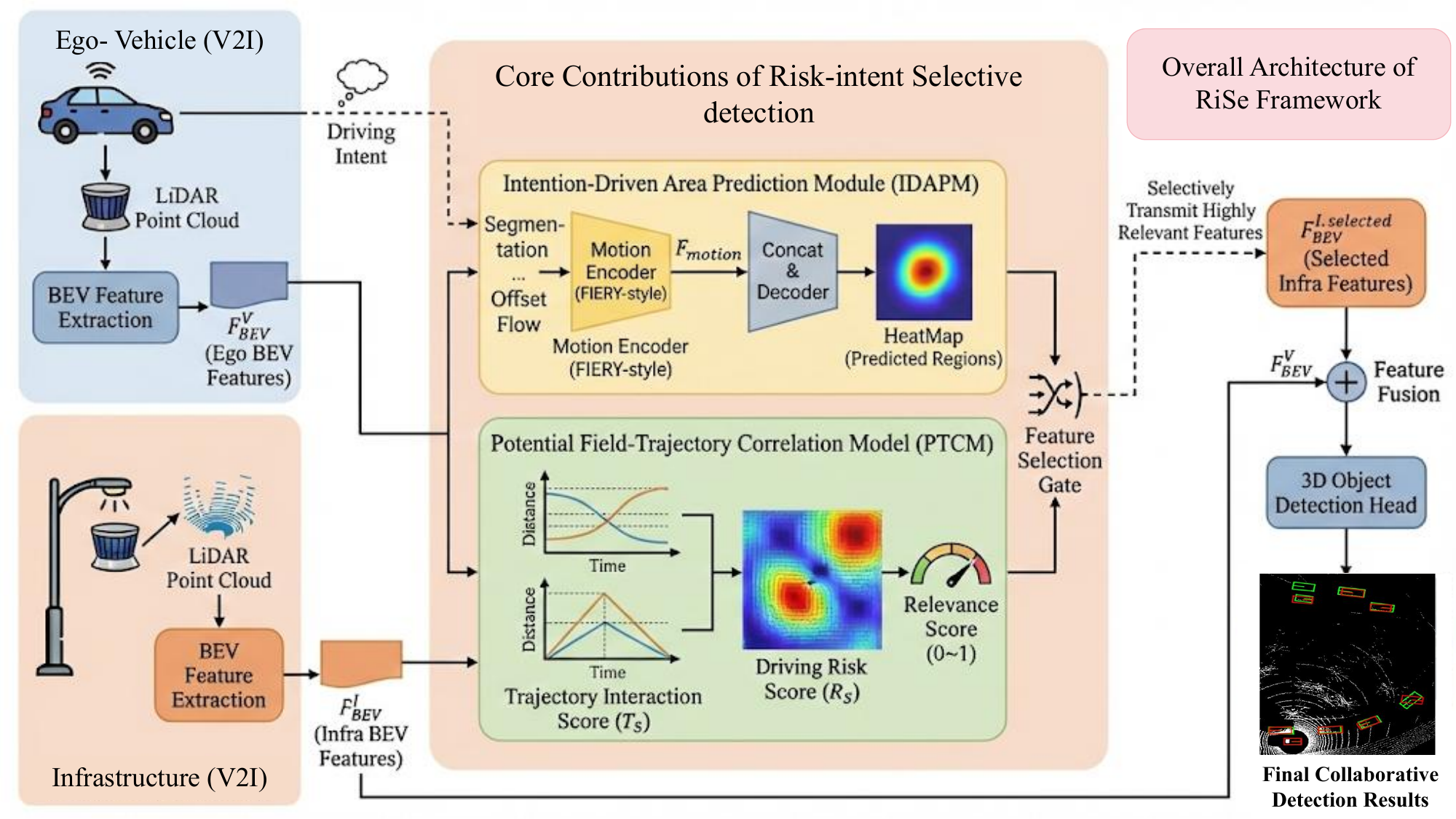}
	\caption[]{The Overall Architecture of RiSe Framework. Guided by the ego-vehicle's driving intent, the infrastructure utilizes the PTCM and IDAPM  to collaboratively generate a correlation heatmap. This heatmap serves as a spatial filter to selectively transmit only high-relevance feature blocks, which are subsequently integrated by the Fusion Module for interaction-aware 3D object detection.}
	\label{fig2}
\end{figure*}

\section{METHOD}
\label{sec:METHOD}
The overall pipeline of RiSe is illustrated in Fig.~\ref{fig2}. We consider a Vehicle-to-Infrastructure (V2I) system where LiDAR point clouds are processed into spatio-temporal BEV features. The core of our framework lies in its intention-driven communication mechanism. Specifically, upon receiving the ego-vehicle's driving intent, the infrastructure activates two key components: the Potential Field-Trajectory Correlation Model (PTCM) and the Intention-Driven Area Prediction Module (IDAPM). As visualized in the architecture, these modules collaboratively assess interaction risks to generate a spatial mask, which filters the raw features. Consequently, only high-relevance feature blocks (depicted in orange) are selectively transmitted to the ego-vehicle. Finally, the vehicle fuses these prioritized features with its local perception stream to perform risk-aware 3D object detection, ensuring that limited bandwidth is strictly allocated to safety-critical regions.

\subsection{Potential Field-Trajectory Correlation Model}
The Potential Field-Trajectory Correlation Model (PTCM) is designed to quantitatively assess target relevance by explicitly coupling future spatial proximity (trajectory interaction) with kinematic consequences (driving risk).\cite{hawlader2025infrastructure} The final relevance score is normalized to a range of 0 to 1, where values closer to 1 indicate higher relevance. The PTCM operates under the rationale that the importance of a traffic participant to the ego-vehicle is jointly determined by their future spatial proximity (trajectory interaction) and the potential consequence of a collision (driving risk). This dual-branch design allows for a more holistic assessment compared to methods relying on a single metric.

\makeatletter
\renewcommand\subsubsection{\@startsection{subsubsection}{3}{\z@}%
  {0pt}%
  {0pt}%
  {\normalfont\normalsize\itshape}}
\makeatother

\renewcommand\thesubsubsection{}
\noindent\subsubsection*{\texorpdfstring{Trajectory Interaction Score.}}\hspace{1em}
This score primarily measures the frequency and extent of trajectory interactions between vehicles. First, the future $N$ frames of motion information (with a time interval $\Delta t$ between frames) are obtained for each autonomous vehicle, serving as the driving trajectory plan at the current moment. Then, the trajectory interaction score is determined by calculating the distances between the cooperative vehicle and other objects at their respective trajectory positions—the closer the distance, the higher the score; the farther the distance, the lower the score. Additionally, to reflect the decreasing importance of future trajectory interactions over time, a gradually diminishing weight is applied to each frame. Where $d$ is the distance in meters, $k$ is the future frame index ($k=1, \dots, N$), and $\lambda_{traj}$ represents the branch weight. $F_S$ is a distance-based potential factor defined by lower and upper thresholds $d_l$ and $d_u$. $W(k)$ applies a decaying weight to each future trajectory frame.

\begin{equation}
F_S=\left\{\begin{matrix}
1 & d \le d_l \\
\frac{d_u - d}{d_u - d_l} & d_l < d < d_u \\
0 & d \ge d_u
\end{matrix}\right..
\end{equation}

\begin{equation}
W(k) = \frac{e^{-k}}{\sum_{j=1}^{N} e^{-j}}.
\end{equation}

\begin{equation}
T_S = \lambda_{traj} \cdot \sum_{k=1}^{N} \left[ W(k) \cdot F_S(k) \right].
\end{equation}

\noindent\subsubsection*{\texorpdfstring{Driving Risk Score.}}\hspace{1em}
This section assesses the potential collision risk between vehicles based on speed information. Due to errors in the speed data from the DeepAccident dataset, which cannot be used directly, we employ a speed differential method to update the target velocities. We compute the ground-relative speed of objects by transforming their positions and velocities from the infrastructure to the world coordinate system. Where $P^{I,t}\left(x,y\right)$ represents the position of the object in the infrastructure coordinate system at time t, ${_I^W}R$ is the rotation matrix from the world coordinate system to the infrastructure coordinate system, and ${{_I^W}T}^t$ is the translation matrix from the world coordinate system to the infrastructure coordinate system at time t. By deriving velocity from position differentials rather than relying on instantaneous sensor readouts, this approach effectively smooths out high-frequency noise and synchronizes the kinematic states between the infrastructure and world frames. Furthermore, in the subsequent potential field construction, we model the risk decay mechanism. The exponential term in the potential energy calculation serves to simulate the 'repulsive force' field characteristic of safety-critical interactions, where the perceived risk escalates non-linearly as the relative distance decreases and the converging velocity increases. This ensures that the model is highly sensitive to imminent collision threats while rapidly ignoring spatially distant and diverging targets.
\begin{equation}
{P^{W,t}\left(x,y\right)=\ {{{_I^W}R}^tP}^{I,t}\left(x,y\right)+{{_I^W}T}^t}.
\end{equation}

\begin{equation}
{V^t(x,y)={{_W^I}R}^t\frac{(P^{W,t}(x,y)-P^{W,t-1}(x,y))}{\triangle t}}.
\end{equation}

Then, a velocity potential field is established based on the speed of the autonomous vehicle. By defining the maximum and minimum ranges for distance and speed, the potential energy is calculated using the vehicle's speed and distance. This value is normalized to a range of 0 to 1 to generate a risk score, reflecting the potential driving risk.

\begin{equation}
{Ev_{ij}=\ \frac{\exp\funcapply(v_{ij}cos\theta_{ij})}{{{(r}_{ij})}^2}}.
\end{equation}
\begin{equation}
{R_S=\ 0.5\cdot\frac{E-E_{min}}{{E_{max}-E}_{min}}}.
\end{equation}


\begin{figure}[!t]
	\centering
	\includegraphics[width=1\linewidth]{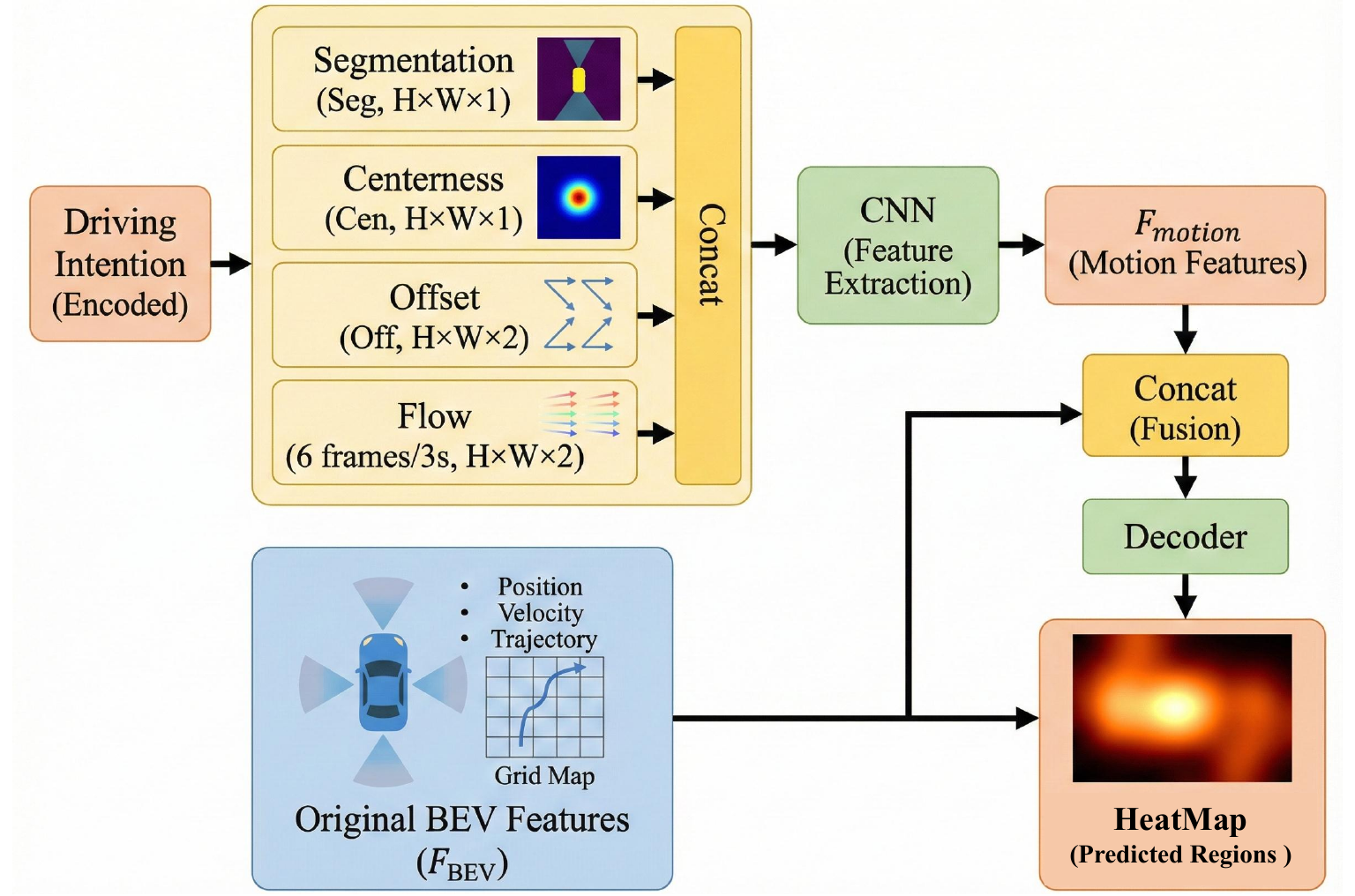}
	\caption{Detailed Architecture of the Intention-Driven Area Prediction Module. The module encodes motion dynamics by extracting four feature components: Segmentation, Centerness, Offset, and Motion Flow. These components are processed by a CNN to generate motion features, which are subsequently fused with the original BEV features to predict the intention-driven relevance heatmap.}
	\label{fig4}
\end{figure}

\subsection{Intention-Driven Area Prediction Module}
The Intention-Driven Area Prediction Module (IDAPM) aims to predict regions relevant to the driving tasks of different autonomous vehicles based on their driving intentions. The architectural diagram is illustrated in Fig.~\ref{fig4}. The original BEV features already include information on the positions, velocities, and trajectories of various objects. However, precise prediction requires explicit identification of the cooperative vehicle. To achieve this, the driving intention of the cooperative vehicle is encoded as a learnable embedding vector and expanded spatially to match the BEV feature dimensions, and the trajectory features are further encoded using the future motion trajectory representation method proposed in e FIERY\cite{gilles2021homeheatmapoutputfuture}. The motion features are then concatenated with the original BEV features, and an output head is attached to the fused features to complete the area prediction. 

\begin{equation}
{HeatMap=Decoder(Concat\left(F_{motion},F_{BEV}\right))}.
\end{equation}

Specifically, we utilize features such as Segmentation, Centerness, Offset, and Flow to model trajectory information. Among these: Segmentation ${\in{\mathbb{R}}}^{H\times W\times1}$ represents the semantic segmentation results. Centerness ${\in{\mathbb{R}}}^{H\times W\times1}$ denotes the probability heatmap of instance centers. Offset ${\in{\mathbb{R}}}^{H\times W\times2}$ describes the offset vectors from the instance center to surrounding points. Flow ${\in{\mathbb{R}}}^{H\times W\times2}$ represents the offset vectors from the current frame's instance center to the next frame's center point. We consider trajectories spanning 6 frames within 3 seconds. After concatenating these features along the feature dimension, a convolutional layer is applied to extract higher-dimensional features. 
\begin{equation}
{F_{motion}=CNN(Concat\left(Seg,Cen,Off,Flow\right)).}
\end{equation}

\subsection{Training Strategy and Loss Function}
Our model employs a two-stage training strategy, and the hyperparameters governing the training process are summarized in Table~\ref{tab:training_hyperparams}. This strategy is adopted to stabilize the learning process and ensure dedicated convergence for each component. The first stage trains the infrastructure model with a composite loss function $\mathcal{L}_{\text{total}}$, while the second stage freezes the infrastructure weights and focuses on optimizing the vehicle-side detection and fusion modules using only $\mathcal{L}_{\text{det}}$. This approach allows the IDAPM module to learn a robust representation of relevant areas based on driving intent before the entire collaborative system is fine-tuned.
\begin{table}[t]
\centering
\caption{Training configuration and hyperparameter specifications for the proposed two-stage collaborative perception framework}
\label{tab:training_hyperparams}
\setlength{\tabcolsep}{12pt}
\begin{tabular}{lc}
\toprule
\textbf{Hyperparameter} & \textbf{Value} \\
\midrule
Optimizer & Adam \\
Learning Rate Schedule & Cosine Annealing \\
Training Strategy & $2$-Stage \\
\midrule
Stage 1 Loss & $\mathcal{L}_{\text{total}} = \mathcal{L}_{\text{det}} + \mathcal{L}_{\text{motion}} + \mathcal{L}_{\text{corr}}$ \\
Stage 2 Loss & $\mathcal{L}_{\text{det}}$ \\
\bottomrule
\end{tabular}
\end{table}
More specifically, in the first stage, the infrastructure model is trained independently, with the loss function consisting of three components: object detection loss $\mathcal{L}_{\text{det}}$, trajectory prediction loss $\mathcal{L}_{\text{motion}}$, and correlated region prediction loss $\mathcal{L}_{\text{corr}}$. The object detection loss utilizes the detection head loss proposed by CenterPoint~\cite{yin2021center}, while the trajectory prediction loss adopts the motion prediction loss introduced by FIERY\cite{gilles2021homeheatmapoutputfuture}. The object detection task focuses on predicting the position and velocity of targets, whereas the trajectory prediction task models the detailed trajectory dynamics of each target. Through the supervision of these tasks, the model effectively learns salient features of each target within the scene. Furthermore, under the ground truth supervision of correlated region heatmaps, it acquires the ability to predict relevant regions based on driving intent. This mechanism explains why our proposed Intention-Driven Area Prediction Module (IDAPM) achieves effective convergence. To further facilitate this module’s convergence, we propose a Rescale Focal Loss. Compared with traditional L1 and L2 loss functions, Rescale Focal Loss preserves the strengths of Focal Loss in managing difficult-to-classify samples, thereby significantly improving prediction accuracy, especially for challenging cases~\cite{chen2023edge}. Standard Focal Loss is originally designed for discrete classification tasks and is suboptimal for our heatmap regression, where the target values are continuous relevance scores ranging from 0 to 1. In our scenario, the `positive' samples (high relevance regions) are extremely sparse compared to the vast background. To address this, we introduce a rescaling factor that dynamically adjusts the loss weight based on the deviation between the predicted interaction score $x$ and the ground truth $gt$. As formulated below, the term $(1 - x/gt)^{\gamma}$ penalizes the model more heavily when it underestimates high-risk regions, while the logarithmic term ensures smooth gradient descent for continuous regression targets:

\begin{equation}
\begin{aligned}
L_{\text{corr}} &= \alpha_{x<gt}\left[-\left(1-\frac{x}{gt}\right)^{\gamma_1}\log\left(\frac{x}{gt}\right)\right] \\
&\quad + \alpha_{x>gt}\left[-\left(1-\frac{1-x}{1-gt}\right)^{\gamma_2}\log\left(\frac{1-x}{1-gt}\right)\right].
\end{aligned}
\end{equation}

In the second stage, we train the complete vehicle-infrastructure collaborative detection model. The infrastructure model’s weights are inherited from the first stage and frozen in this phase, ensuring a dedicated focus on the intention-driven correlated region inference task. Meanwhile, the full model focuses on learning the weights of the vehicle-side object detection and feature fusion modules. The loss function in this stage includes only the object detection loss. To maintain training stability and efficiency, we employ a cosine annealing learning rate schedule and utilize the Adam optimizer across both stages, with training conducted over a total of 20 epochs.

\begin{table*}[]
\small
\centering
\caption{\textcolor{black}{Comparative experiments on the \textbf{3D object detection benchmark}. In order to ensure fairness, all methods are based on PointPillars\cite{lang2019pointpillars} as the baseline, using the same fusion module while only changing the features transmitted during communication.}}
\label{table_1} 
\renewcommand\arraystretch{1.8}
\setlength{\tabcolsep}{4pt}
\begin{tabularx}{\textwidth}{r|c|XXXXXXX|c}
\toprule
\textbf{Method} & \textbf{Fusion Type} & \textbf{Car} & \textbf{Van} & \textbf{Truck} & \textbf{Cyclist} & \textbf{Motorcycle} & \textbf{Pedestrian} & \textbf{Average} & \textbf{Comm. Vol} \\ 
\midrule
PointPillars\cite{lang2019pointpillars} & Veh.-Only & 37.01 & 23.68 & 26.41 & 17.42 & 29.04 & 20.76 & 25.72 & 0 \\ 
PointPillars\cite{lang2019pointpillars} & Intermediate & 44.10 & 25.68 & 28.63 & 16.43 & 25.04 & 30.86 & 28.46 & 100\% \\ 
When2Com\cite{liu2020when2commultiagentperceptioncommunication} & Intermediate & 42.49 & 24.16 & 31.33 & 13.49 & 29.58 & 26.82 & 27.98 & 54.18\% \\ 
Where2Comm\cite{hu2022where2commcommunicationefficientcollaborativeperception} & Intermediate & 44.25 & 26.96 & 32.25 & 16.86 & 27.04 & 33.78 & 30.19 & 1.19\% \\ 
V2VNet\cite{wang202022vnetvehicletovehiclecommunicationjoint} & Intermediate & 38.97 & 29.54 & 32.63 & 15.64 & 30.96 & 29.21 & 29.49 & 0.24\% \\ 
V2X-ViT\cite{xu2022v2xvitvehicletoeverythingcooperativeperception} & Intermediate & 39.79 & 26.30 & 30.75 & 20.83 & 30.21 & 30.77 & 29.77 & 3.13\% \\ 
\rowcolor{mygreen} RiSe (Ours) & Intermediate & 49.26 & 29.77 & 38.22 & 20.37 & 33.06 & 29.63 & 33.39 & 0.71\% \\ 
\bottomrule
\end{tabularx}
\end{table*}

\section{EXPERIMENTS}
\label{sec:EXPERIMENTS}
\subsection{Dataset and Implementation Details}
We evaluate RiSe on the DeepAccident dataset~\cite{wang2024deepaccident}, which contains 57K annotated frames focused on safety-critical scenarios. We follow the standard split (0.7/0.15/0.15) and detection range ($[-51.2\text{m}, 51.2\text{m}]$ for $X/Y$ axes) with a voxel resolution of $0.8\text{m}$.
The feature extraction backbone is based on PointPillars~\cite{lang2019pointpillars}.
Models are trained for 20 epochs using the Adam optimizer (initial $lr=10^{-3}$, cosine annealing) on 8 NVIDIA RTX 4090 GPUs.
Regarding hyperparameters, we set $\alpha=1.0, \gamma=2.0$ for Rescale Focal Loss, and $d_l=5\text{m}, d_u=20\text{m}, \lambda_{traj}=0.5$ for the interaction model.
For evaluation, we report the standard mean Average Precision (mAP) and Communication Volume (log base 2). Additionally, we introduce \textit{corr-mIoU} and \textit{IoU-error} to assess the precision of intention-driven region prediction. All baselines are re-implemented under the same settings for fair comparison.

\subsection{Main Results}
Table~\ref{table_1} presents a comprehensive comparison of detection accuracy and communication overhead. 
While standard Intermediate Fusion (Full) increases the mAP from the single-vehicle baseline of $25.72\%$ to $28.46\%$, it imposes a prohibitive 100\% bandwidth load. 
Compression-based approaches mitigate this but often at the cost of precision. 
Notably, V2VNet~\cite{wang202022vnetvehicletovehiclecommunicationjoint} reduces volume to $0.24\%$ but suffers significant information loss, causing the AP for Cars to drop to $38.97\%$, which is inferior even to the uncompressed baseline. 
Similarly, V2X-ViT~\cite{xu2022v2xvitvehicletoeverythingcooperativeperception} achieves only $29.77\%$ mAP despite utilizing sophisticated transformers.

In contrast, RiSe establishes a superior Pareto frontier. 
With a communication volume of merely $0.71\%$, our method achieves an mAP of 33.39\%, outperforming both the full-communication baseline and the state-of-the-art method Where2Comm~\cite{hu2022where2commcommunicationefficientcollaborativeperception} ($30.19\%$). 
Crucially, our Interaction-Aware Mechanism ensures the preservation of safety-critical features. 
This is evidenced by our performance on high-risk categories: we secure the highest AP for Cars (49.26\%) and Trucks (38.22\%) across all collaborative methods. 
These results validate that intent-driven semantic filtering effectively distinguishes risk-critical targets from background redundancy, offering a more robust solution than indiscriminate compression.

\subsection{Ablation Study}
\textbf{Impact of PTCM.} To verify the efficacy of the dual-branch architecture, we compare the complete PTCM against single-branch variants (TS-only, RS-only) and the Full-communication baseline. 
As detailed in Table~\ref{tab:ptcm_ablation}, the full PTCM yields the optimal trade-off, achieving $33.39\%$ mAP and $58.37\%$ Corr-mIoU with merely $0.71\%$ communication volume.
Isolating the Trajectory Score (TS-only) results in a performance drop to $30.19\%$ mAP and a bandwidth increase to $0.95\%$, underscoring the necessity of velocity-based risk modeling for precise filtering.
Conversely, the RS-only variant suffers a more severe degradation ($29.44\%$ mAP), confirming that trajectory interaction provides indispensable spatiotemporal context.
Compared to the Full-communication baseline ($28.46\%$ mAP, $100\%$ Comm.), our complete model significantly boosts accuracy while minimizing bandwidth, conclusively validating the complementary nature of kinematic risk and trajectory interaction.
\begin{table}[!ht]
\caption{Ablation Study on PTCM Components: Impact of Trajectory Score (TS) and Risk Score (RS) Branches.}
\label{tab:ptcm_ablation}
\centering
\small
\renewcommand{\arraystretch}{1.4}
\setlength{\tabcolsep}{3pt}
\begin{tabular}{l|ccc}
\toprule
\textbf{Method} & \textbf{Comm. Vol} $\downarrow$ & \textbf{mAP} $\uparrow$ & \textbf{Corr-mIoU} $\uparrow$ \\
\midrule
Baseline (Full Comm.) & 100\% & 28.46 & 55.33 \\
TS only (No RS) & 0.95\% & 30.19 & 57.37 \\
RS only (No TS) & 0.88\% & 29.44 & 57.54 \\
\rowcolor{mygreen} Full PTCM (TS+RS) & 0.71\% & 33.39 & 58.37 \\
\bottomrule
\end{tabular}
\vspace{0.2cm}
\end{table}

\textbf{Impact of IDAPM.}
The Intention-Driven Area Prediction Module (IDAPM) enhances correlation predictions by exploiting temporal, motion, and velocity cues. We evaluate its impact using \textit{corr-mIoU} (measuring overlap quality) and \textit{IoU-error} (penalizing redundancy).

\begin{table}[!t] 
\small
\centering
\renewcommand\arraystretch{1.5}
\setlength{\tabcolsep}{8pt}

\caption{Effect of Temporal Information. Temporal cues improve overall accuracy and correlation prediction.}
\label{table_2a}
\begin{tabular}{c|ccc}
\hline
\textbf{Temporal} & \textbf{mAP} $\uparrow$ & \textbf{corr-mIoU} $\uparrow$ & \textbf{IoU-error} $\downarrow$ \\ 
\hline
$ \times $  & 28.62 & 55.33 & 11.76 \\ 
\hline
\rowcolor{mygreen}$ \checkmark $ & 32.45 & 57.37 & 14.96 \\
\hline
\textit{Improvement} & \textbf{\textcolor{textgreen}{+3.83}} & \textbf{\textcolor{textgreen}{+2.04}} & \textbf{\textcolor{textgreen}{+3.20}} \\
\hline
\end{tabular}

\vspace{10pt} 

\caption{Effect of Motion Supervision. Motion supervision enhances correlation reasoning in dynamic scenes.}
\label{table_2b}
\begin{tabular}{c|ccc}
\hline
\textbf{Motion} & \textbf{mAP} $\uparrow$ & \textbf{corr-mIoU} $\uparrow$ & \textbf{IoU-error} $\downarrow$ \\ 
\hline
$ \times $ & 32.45 & 57.37 & 14.96 \\ 
\hline
\rowcolor{mygreen}$ \checkmark $ & 33.01 & 57.54 & 12.86 \\
\hline
\textit{Improvement} & \textbf{\textcolor{textgreen}{+0.56}} & \textbf{\textcolor{textgreen}{+0.17}} & \textbf{\textcolor{textgreen}{-2.10}} \\
\hline
\end{tabular}

\vspace{10pt} 

\caption{Effect of Velocity Supervision. Velocity supervision provides quantitative speed cues and reduces redundancy.}
\label{table_2c}
\begin{tabular}{c|ccc}
\hline
\textbf{Velocity} & \textbf{mAP} $\uparrow$ & \textbf{corr-mIoU} $\uparrow$ & \textbf{IoU-error} $\downarrow$ \\ 
\hline
$ \times $ & 33.01 & 57.54 & 12.86 \\ 
\hline
\rowcolor{mygreen}$ \checkmark $ & 33.39 & 58.37 & 11.18 \\
\hline
\textit{Improvement} & \textbf{\textcolor{textgreen}{+0.38}} & \textbf{\textcolor{textgreen}{+0.83}} & \textbf{\textcolor{textgreen}{-1.68}} \\
\hline
\end{tabular}

\end{table}

\textbf{Effect of Temporal Information.}
As shown in Table~\ref{table_2a}, incorporating temporal cues yields a notable performance boost. The model achieves an mAP of $32.45\%$, representing a substantial gain of $3.83\%$ over the single-frame baseline. Similarly, correlation accuracy improves to $57.37\%$, confirming that temporal continuity is pivotal for stabilizing trajectory regression.

\textbf{Effect of Motion Supervision.}
Table~\ref{table_2b} highlights the benefits for dynamic agents. While mAP rises to $33.01\%$, the significant reduction in IoU-error (from $14.96\%$ to $12.86\%$) is most critical. This indicates that motion supervision effectively suppresses false positives, allowing the network to better distinguish moving targets from the background.

\textbf{Effect of Velocity Supervision.}
Finally, as demonstrated in Table~\ref{table_2c}, integrating velocity supervision provides necessary constraints for risk assessment. This addition pushes mAP to $33.39\%$ and reduces IoU-error to its lowest point ($11.18\%$), validating that velocity-aware loss minimizes redundant predictions and strengthens dynamic risk modeling.

\textbf{Impact of Rescale Focal Loss.} 
Table~\ref{table_3} compares the proposed objective against standard baselines (L1, L2, Focal Loss) on correlation prediction precision. 
Our Rescale Focal Loss achieves a superior \textit{corr-mIoU} of $58.37\%$, significantly outperforming the standard Focal Loss ($48.26\%$). 
This substantial margin suggests that the rescaling mechanism successfully adapts the hard-mining capability of classification losses to the regression task, thereby ensuring more precise localization of intention-driven regions.
\begin{table}[t]
\small
\centering
\caption{Effect of Rescale Focal Loss on IDAPM's prediction quality. The
evaluation highlights the competitive performance of our loss in target region
localization.}
\label{table_3}
\renewcommand\arraystretch{1.3}
\setlength{\tabcolsep}{6pt}
\begin{tabular}{c|cc}
\hline
\textbf{Loss Function} & \textbf{corr-mIoU} $\uparrow$ & \textbf{IoU-error} $\downarrow$ \\
\hline
Focal Loss & 48.26 & 38.31 \\
L1 Loss & 39.42 & 49.50 \\
L2 Loss & 49.58 & 10.00 \\
\rowcolor{mygreen} \textbf{Rescale Focal Loss (Ours)} & \textbf{58.37} & \textbf{11.18} \\
\hline
\end{tabular}
\end{table}

\subsection{Visualization}
In this section, we visualized the 3D object detection results and the correlation prediction results.
\begin{figure}[!h]
	\centering
	\includegraphics[width=\linewidth]{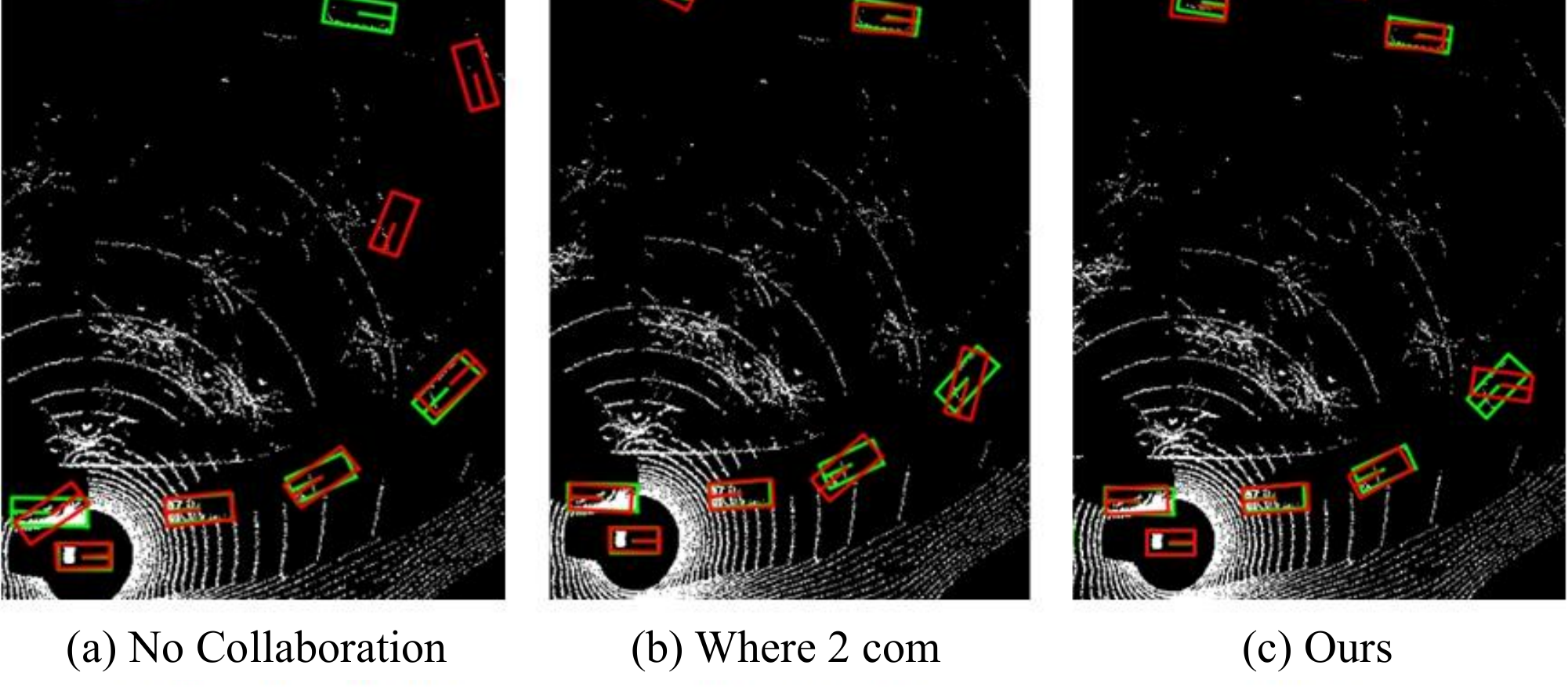}
	\caption{BEV Visualization of 3D Object Detection. The first image indicates a 3D object detection result without collaborative communication. The middle shows the result from the Where2comm method. The last shows the 3D detection result with the method proposed in this paper.}
	\label{fig6}
\end{figure}

\textbf{Qualitative Results at Intersection. }As shown in  Fig.~\ref{fig6}, this is a typical intersection scenario. The green boxes represent the ground truth, while the red boxes represent the predicted results. Due to sensor occlusion, the vehicle above the intersection could not be detected without collaboration, as there are no point clouds available in that region. However, with vehicle-infrastructure collaboration, the vehicle above the intersection was successfully detected, demonstrating the ability of collaboration to effectively expand the perception range and overcome sensor occlusion. Furthermore, as illustrated in the subsequent two figures, compared to the Where2Comm\cite{hu2022where2commcommunicationefficientcollaborativeperception} algorithm, our approach achieves comparable detection performance while significantly reducing communication bandwidth requirements. This not only underscores the efficiency of our method but also highlights its efficacy in balancing detection accuracy and communication cost, making it a more practical solution for real-world applications.
\begin{figure}[!h]
	\centering
	\includegraphics[width=\linewidth]{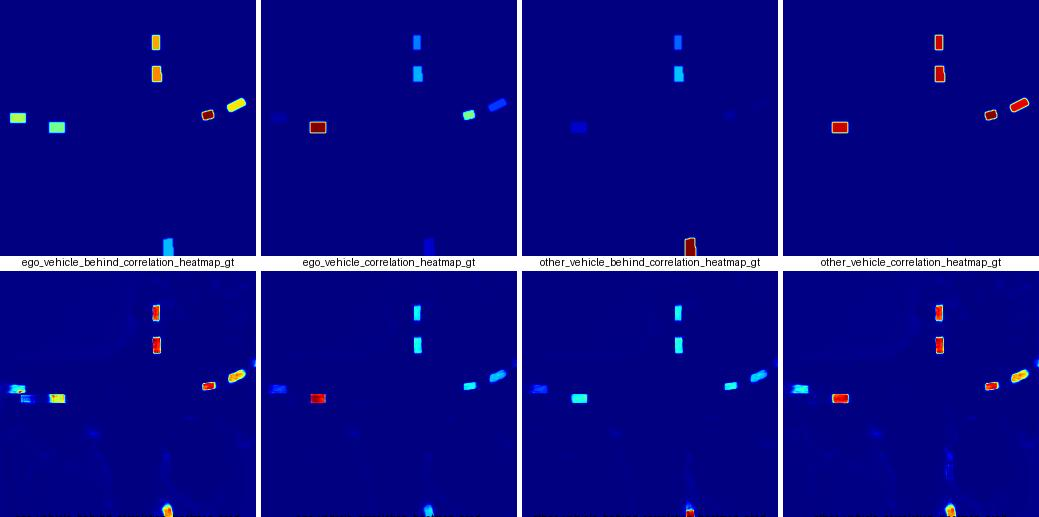}
	\caption{Visualization of Intention-Aware Correlation Prediction. The top row displays the ground truth heatmaps, while the bottom row presents the predicted results. The color spectrum represents the relevance score: warmer colors indicate high correlation and interaction risk, whereas cooler colors denote low-relevance regions. The figure demonstrates that the infrastructure system can accurately predict correlations based on vehicle intentions and interaction risks.}
	\label{fig7}
\end{figure}
\textbf{Intention-Aware Correlation Prediction. }In Fig.~\ref{fig7}, the top section illustrates the ground truth (GT), while the bottom section presents the predicted results. The visualization demonstrates that when different vehicles transmit their respective driving intentions to the infrastructure, the system can effectively predict correlations by analyzing these intentions and their interaction risks with other vehicles. For instance, the trajectory of the vehicle on the left moves from right to left, remaining relatively close to the upper vehicle and showing frequent trajectory interactions. Consequently, its correlation score is higher. In contrast, the trajectory of the vehicle on the right moves from left to right, maintaining a greater distance from the upper vehicle. This results in a lower correlation score, aligning well with expectations. This consistency further validates the effectiveness of the proposed model in capturing intention-based interactions.

The visualizations confirm that the proposed correlation-aware framework not only reduces communication overhead but also ensures accurate and consistent detection performance under different traffic conditions.

\section{LIMITATION AND DISCUSSION}
\label{sec:LIMITATION AND DISCUSSION}
While RiSe establishes an efficient trade-off between bandwidth and accuracy, several limitations persist. First, the current framework is LiDAR-centric and vehicle-focused; extending it to multi-modal and multi-class perception would enhance versatility. Second, the computational overhead of PTCM and IDAPM constrains deployment on low-power edge platforms, necessitating future exploration of lightweight architectures or model distillation. Third, the assumption of synchronized communication simplifies real-world complexities, highlighting the need for fusion mechanisms robust to latency and asynchrony. Additionally, IDAPM relies on provided intent signals; practical deployment requires integrating real-time intention estimation, potentially through joint optimization to mitigate errors. Finally, the issue of data privacy in collaborative transmission is not explicitly addressed; future work could explore privacy-preserving mechanisms to secure shared feature streams against potential information leakage in open V2X networks.

\section{CONCLUSION}
\label{sec:CONCLUSION}
This paper presents RiSe, an interaction-aware V2I framework that resolves the conflict between bandwidth constraints and detection performance by prioritizing information based on interaction criticality. By integrating the Potential Field-Trajectory Correlation Model (PTCM) to quantify kinematic risks and the Intention-Driven Area Prediction Module (IDAPM) to filter redundant features using ego-motion priors, the system optimizes data transmission. Experiments on the DeepAccident dataset demonstrate an effective Pareto frontier, achieving state-of-the-art accuracy for safety-critical categories while reducing communication volume to 0.71\% of full transmission. Additionally, the proposed Rescale Focal Loss improves high-risk target localization in complex scenarios. Future work will address multi-modal fusion, asynchronous communication, real-time intention estimation, and data privacy mechanisms.

\bibliographystyle{Transactions-Bibliography/IEEEtran}
\bibliography{Transactions-Bibliography/IEEEabrv,Transactions-Bibliography/reference}

\end{CJK}
\end{document}